# K-ANMI: A Mutual Information Based Clustering Algorithm for Categorical Data

Zengyou He, Xiaofei Xu, Shengchun Deng

*Department of Computer Science and Engineering, Harbin Institute of Technology,*

*92 West Dazhi Street, P.O Box 315, P. R. China, 150001*

zengyouhe@yahoo.com, {xiaofei, dsc}@hit.edu.cn

**Abstract** Clustering categorical data is an integral part of data mining and has attracted much attention recently. In this paper, we present $k$-ANMI, a new efficient algorithm for clustering categorical data. The $k$-ANMI algorithm works in a way that is similar to the popular $k$-means algorithm, and the goodness of clustering in each step is evaluated using a mutual information based criterion (namely, Average Normalized Mutual Information-*ANMI*) borrowed from cluster ensemble. Experimental results on real datasets show that $k$-ANMI algorithm is competitive with those state-of-art categorical data clustering algorithms with respect to clustering accuracy.

## 1. Introduction

Clustering is an important data mining technique that groups together similar data records. Recently, much attention has been put on clustering categorical data [e.g., 1-21, 26 -31], where records are made up of non-numerical attributes. Fast and accurate clustering of categorical data has many potential applications in customer relationship management, e-commerce intelligence, etc.

In [21], the categorical data clustering problem is defined as an optimization problem using a mutual information sharing based object function (namely, Average Normalized Mutual Information-*ANMI*) from the viewpoint of cluster ensemble. However, those algorithms in [21] have been developed from intuitive heuristics rather than from the vantage point of a direct optimization, which can't guarantee to find a reasonable solution.

In this paper, we propose a new $k$-means like clustering algorithm called $k$-ANMI for categorical data that directly optimizes the mutual information sharing based object function. The $k$-ANMI algorithm takes the number of desired clusters (supposed to be $k$) as input and iteratively changes the class label of each data object to improve the value of object function. That is, for each object, the current label is changed to each of the other $k − 1$ possible labels and the ANMI objective is re-evaluated. If the ANMI increases, the object's label is changed to the best new value and the algorithm proceeds to the next object. When all objects have been checked for possible improvements, a sweep is completed. If at least one label was changed in a sweep, we initiate a new sweep. The algorithm terminates when a full sweep does not change any labels, thereby indicating that a local optimum is reached.

Although the basic idea of $k$-ANMI is very simple, it is nontrivial to effectively implement the algorithm so that it is scalable to large datasets. To make the $k$-ANMI algorithm scalable, we employ multiple hash tables to improve its efficiency. We also provide the analysis on the time complexity and space complexity of $k$-ANMI algorithm. The analysis shows that the $k$-ANMI algorithm is capable of handing large categorical datasets.

The remainder of this paper is organized as follows. Section 2 presents a critical review on related work. Section 3 introduces basic concepts and formulates the problem. In Section 4, we present the $k$-ANMI algorithm and provide complexity analysis. Experimental results are given in Section 5 and Section 6 concludes the paper.

## 2. Related Work

A few algorithms have been proposed in recent years for clustering categorical data [1-2,26 3]. In [1], the problem of clustering customer transactions in a market database is addressed. STIRR, an iterative algorithm based on non-linear dynamical systems is presented in [2]. The approach used in [2] can be mapped to a certain type of non-linear systems. If the dynamical system converges, the categorical databases can be clustered. Another recent research [3] shows that the known dynamical systems cannot guarantee convergence, and proposes a revised dynamical system in which convergence can be guaranteed.

K-modes, an algorithm extending the $k$-means paradigm to categorical domain is introduced in [4,5]. New dissimilarity measure to deal with categorical data is conducted to replace means with modes, and a frequency based method is used to update modes in the clustering process to minimize the clustering cost function. Based on $k$-modes algorithm, [6] proposes an adapted mixture model for categorical data, which gives a probabilistic interpretation of the criterion optimized by the $k$-modes algorithm. A fuzzy $k$-modes algorithm is presented in [7] and tabu search technique is applied in [8] to improve fuzzy $k$-modes algorithm. An iterative initial-points refinement algorithm for categorical data is presented in [9]. The work in [19] can be considered as an extension of $k$-modes algorithm to transaction domain.

In [10], the authors introduce a novel formalization of a cluster for categorical data by generalizing a definition of cluster for numerical data. A fast summarization based algorithm, CACTUS, is presented. CACTUS consists of three phases: *summarization*, *clustering*, and *validation*.

ROCK, an adaptation of an agglomerative hierarchical clustering algorithm, is introduced in [11]. This algorithm starts by assigning each tuple to a separated cluster, and then clusters are merged repeatedly according to the closeness between clusters. The closeness between clusters is defined as the sum of the number of "links" between all pairs of tuples, where the number of "links" is computed as the number of common neighbors between two tuples.

In [12], the authors propose the notion of *large item*. An item is *large* in a cluster of transactions if it is contained in a user specified fraction of transactions in that cluster. An allocation and refinement strategy, which has been adopted in partitioning algorithms such as $k$-means, is used to cluster transactions by minimizing the criteria function defined with the notion of large item. Following the large item method in [12], a new measurement, called the small-large ratio is proposed and utilized to perform the clustering [13]. In [14], the authors consider the item taxonomy in performing cluster analysis. Xu and Sung [15] propose an algorithm based on "caucus", which is fine-partitioned demographic groups that is based the purchase features of customers.

Squeezer, a one-pass algorithm is proposed in [16]. *Squeezer* repeatedly read tuples from dataset one by one. When the first tuple arrives, it forms a cluster alone. The consequent tuples are either put into an existing cluster or rejected by all existing clusters to form a new cluster according to the given similarity function.

COOLCAT, an entropy-based algorithm for categorical clustering, is proposed in [17]. Based on height-to-width ratio of the cluster histogram, Yang et al. [18] develop the CLOPE algorithm. Ref. [20] introduces a distance measure between partitions based on the notion of generalized conditional entropy and a genetic algorithm approach is utilized for discovering the median partition. LIMBO introduced in [27] is a scalable hierarchical categorical clustering algorithm that builds on the Information Bottleneck framework. Li et al. [28] shows that the entropy-based criterion in categorical data clustering can be derived in the formal framework of probabilistic clustering models and develops an iterative Monte-Carlo procedure to search for the partitions minimizing the criterion.

In [21], the authors formally define the categorical data clustering problem as an optimization problem from the viewpoint of cluster ensemble, and apply cluster ensemble approach for clustering categorical data. Simultaneously, Gionis et al. [30] use cluster ensemble methods with disagreement measure to solve the problem of categorical data clustering.

Chen and Chuang [26] develop the CORE algorithm by employing the concept of correlated-force ensemble. He et al. [29] propose TCSOM algorithm for clustering binary data by extending traditional self-organizing map (SOM). Chang and Ding [31] present a method for visualization of the clustered categorical data such that users' subjective factors can be reflected by adjusting clustering parameters, and therefore to increase the clustering result's reliability.

## 3. Introductory Concepts and Problem Formulation

### 3.1 Notations

Let $A_1, ..., A_r$ be a set of categorical attributes with domains $D_1,..., D_r$ respectively. Let the dataset $D = \{X_1, X_2, ..., X_n\}$ be a set of objects described by $r$ categorical attributes, $A_1, ..., A_r$. The value set $V_i$ of $A_i$ is set of values of $A_i$ that are present in $D$. For each $v \in V_i$, the frequency $f(v)$, denoted as $f_v$, is number of objects $O \in X$ with $O.A_i = v$. Suppose the number of distinct attribute values of $A_i$ is $p_i$, we define the histogram of $A_i$ as the set of pairs: $h_i = \{(v_1, f_1), (v_2, f_2),..., (v_{p_i}, f_{p_i})\}$ and the size of $h_i$ is $p_i$. The histograms of the data set $D$ is defined as: $H = \{h_1, h_2, ..., h_r\}$.

Let $X, Y$ be two categorical objects described by $r$ categorical attributes. The dissimilarity measure between $X$ and $Y$ can be defined by the total mismatches of the corresponding attribute values of the two objects. The smaller the number of mismatches is, the more similar the two objects. Formally,

$$d_1(X,Y) = \sum_{j=1}^{r} \delta(x_j, y_j) \tag{1}$$

where

$$\delta(x_j, y_j) = \begin{cases} 0 & (x_j = y_j) \\ 1 & (x_j \neq y_j) \end{cases} \quad (2)$$

Given the dataset $D = \{X_1, X_2, …, X_n\}$ and an object $Y$, The dissimilarity measure between $X$ and $Y$ can be defined by the average of the sum of the distances between $X_i$ and $Y$.

$$d_2(D, Y) = \frac{\sum_{j=1}^{n} d_1(X_j, Y)}{n} \quad (3)$$

If we take the histograms $H = \{h_1, h_2, …, h_m\}$ as the compact representation of the data set $D$, formula (3) can be redefined as (4).

$$d_3(H, Y) = \frac{\sum_{j=1}^{r} \phi(h_j, y_j)}{n} \quad (4)$$

where

$$\phi(h_j, y_j) = \sum_{l=1}^{p_j} f_l * \delta(v_l, y_j) \quad (5)$$

From a viewpoint of implementation efficiency, formula (4) can be presented in form of (6).

$$d_4(H, Y) = \frac{\sum_{j=1}^{r} \psi(h_j, y_j)}{n} \quad (6)$$

where

$$\psi(h_j, y_j) = \sum_{l=1}^{p_j} f_l * (1 - \delta(v_l, y_j)) \quad (7)$$

Formula (6) can be computed more efficiently for which requires only the frequencies of matched attribute value pairs.

### 3.2 A Unified View in the Cluster Ensemble Framework

Cluster ensemble (CE) is the method to combine several runs of different clustering algorithms to get a common partition of the original dataset, aiming for consolidation of results from a portfolio of individual clustering results.

Clustering aims at discovering groups and identifying interesting patterns in a dataset. We call a particular clustering algorithm with a specific view of the data a *clusterer*. Each clusterer outputs a *clustering* or *labeling*, comprising the group labels for some or all objects.

Given dataset $D = \{X_1, X_2, …, X_n\}$, a partitioning of these $n$ objects into $k$ clusters can be represented as a set of $k$ sets of objects $C_l = \{l=1, …, k\}$ or as a label vector $\lambda \in N^n$. A clusterer $\Phi$ is a function that delivers a label vector given a set of objects. Fig.1 (adapted from [24]) shows the basic setup of the cluster ensemble: A set of $r$ labelings $\lambda^{(1,2,...,r)}$ is combined into a single labeling $\lambda$ (the *consensus labeling*) using a consensus function $\Gamma$.

As shown in [21], categorical data clustering problem can be considered as the cluster ensemble problem. That is, for a categorical dataset, if we consider attribute values as cluster labels, each attribute with its attribute values give a "*best clustering*" on the dataset without considering other attributes. So the categorical data clustering problem can be considered as the cluster ensemble problem, in which the attribute values of each attribute are the outputs of different clustering algorithms.

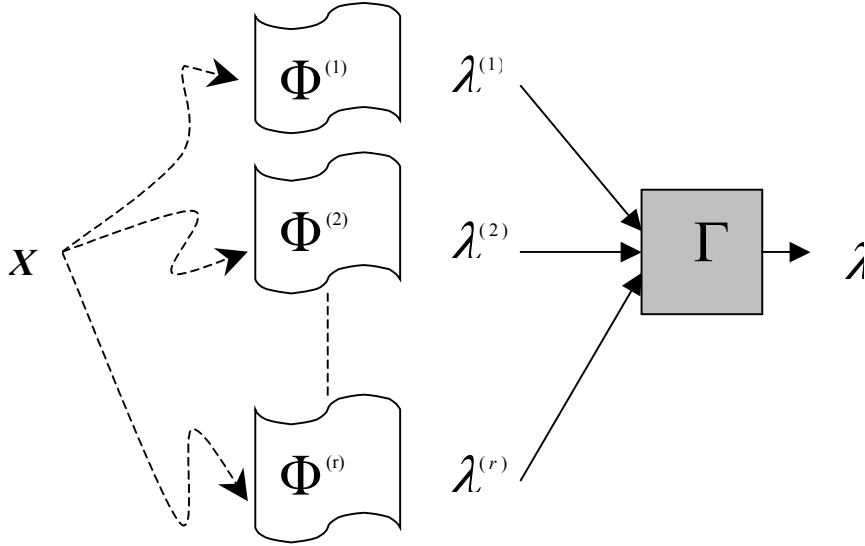

Fig. 1. The cluster ensemble. A consensus function $\Gamma$ combines clusterings $\lambda^{(q)}$ from a variety of sources.

More precisely, for the dataset $D = \{X_1, X_2, ..., X_n\}$ with $r$ categorical attributes, $A_1, ..., A_r$. The value set $V_i$ is a set of values of $A_i$ that are present in $D$. According to the CE framework described in Fig.1, if we define each clusterer $\Phi^{(i)}$ as a function that mapping values in $V_i$ to distinct natural numbers, we can get the optimal partitioning $\lambda^{(i)}$ determined by each attribute $A_i$ as: $\lambda^{(i)} = \{\Phi^{(i)}(X_j.A_i) \mid X_j.A_i \in V_i, X_j \in D\}$. So, we can combine the set of $r$ labelings $\lambda^{(1,2,...,r)}$ into a single labeling $\lambda$ using a consensus function $\Gamma$ to get the solution for the problem of clustering categorical data.

### 3.3 Object Function

Continue Section 3.2, intuitively, a good combined clustering should share as much information as possible with the given $r$ labelings. Strehl and Ghosh [22-24] use the mutual information in information theory to measure the shared information, which can be directly applied in this literature.

More concisely, as shown in Strehl's papers [23,24], given $r$ groupings with the $q$-th grouping $\lambda^{(q)}$ having $k^{(q)}$ clusters, a consensus function $\Gamma$ is defined as a function $N^{n \times r} \to N^n$ mapping a set of clusterings to an integrated clustering:

$$\Gamma : \{\lambda^{(q)} \mid q \in \{1,2,...,r\}\} \to \lambda \tag{8}$$

The set of groupings is denoted as $\Lambda = \{\lambda^{(q)} \mid q \in \{1,2,...,r\}\}$. The optimal combined clustering should share the most information with the original clusterings. In information theory, mutual information is a symmetric measure to quantify the statistical information shared between two distributions. Let $A$ and $B$ be the random variables described by the cluster labeling $\lambda^{(a)}$ and $\lambda^{(b)}$, with $k^{(a)}$ and $k^{(b)}$ gruops respectively. Let $I(A, B)$ denote the mutual information between $A$ and $B$, and $H(A)$ denote the entropy of $A$. As Strehl has shown in [24], $I(A,B) \leq \frac{H(A)+H(B)}{2}$ holds. Hence, the [0,1]-normalized mutual information (NMI) [24] used is:

$$NMI(A,B) = \frac{2I(A,B)}{H(A)+H(B)} \tag{9}$$

Obviously, $NMI(A, A) = 1$. Equation (9) has to be estimated by the sampled quantities provided by the clusterings [24]. As shown in [24], if we let $n^{(h)}$ be the number of objects in cluster $C_h$ according to $\lambda^{(a)}$, and let $n_g$ be the number of objects in cluster $C_g$ according to $\lambda^{(b)}$. Let $n_g^{(h)}$ be denote the number of objects in cluster $C_h$ according to $\lambda^{(a)}$ as well as in cluster $C_g$ according to $\lambda^{(b)}$. The [0,1]-normalized mutual information criteria $\phi^{(NMI)}$ is computed as follows [23,24]:

$$\phi^{(NMI)}(\lambda^{(a)}, \lambda^{(b)}) = \frac{2}{n} \sum_{h=1}^{k^{(a)}} \sum_{g=1}^{k^{(b)}} n_g^{(h)} \log_{k^{(a)} * k^{(b)}} (\frac{n_g^{(h)} n}{n^{(h)} n_g}) \tag{10}$$

Therefore, the Average Normalized Mutual Information (ANMI) between a set of $r$ labelings, $\Lambda$, and a labeling $\tilde{\lambda}$ is defined as follows [24]:

$$\phi^{(ANMI)}(\Lambda, \tilde{\lambda}) = \frac{1}{r} \sum_{q=1}^{r} \phi^{(NMI)}(\tilde{\lambda}, \lambda^{(q)}) \tag{11}$$

According to [23,24], the optimal combined clustering $\lambda^{(k-opt)}$ should be defined as the one that has the maximal average mutual information with all individual partitioning $\lambda^{(q)}$ given that the number of consensus clusters desired is *k*. Thus the objective function for categorical data clustering is Average Normalized Mutual Information (*ANMI*). Then, $\lambda^{(k-opt)}$ is defined as [24]:

$$\lambda^{(k-opt)} = \arg\max_{\tilde{\lambda}} \sum_{q=1}^{r} \phi^{(NMI)}(\tilde{\lambda}, \lambda^{(q)}) \tag{12}$$

where $\tilde{\lambda}$ goes through all possible *k*-partitions.

Taking *ANMI* as the object function in our *k*-ANMI algorithm, we have to compute the value of $\phi^{(NMI)}$. More precisely, we should be able to efficiently get the accurate value of each parameter in Equation (10). In the next Section, we will describe our data structures and *k*-ANMI algorithm in detail.

## 4. The *k*-ANMI Algorithm

### 4.1 Overview

The *k*-ANMI algorithm takes the number of desired clusters (supposed to be *k*) as input and iteratively changes the class label of each data object to improve the value of object function. That is, for each object, the current label is changed to each of the other *k* − 1 possible labels and the ANMI objective is re-evaluated. If the ANMI increases, the object's label is changed to the best new value and the algorithm proceeds to the next object. When all objects have been checked for possible improvements, a sweep is completed. If at least one label was changed in a sweep, we initiate a new sweep. The algorithm terminates when a full sweep does not change any labels, thereby indicating that a local optimum is reached.

### 4.2 Data Structures

Taking the dataset ***D*** = {$X_1, X_2, \ldots, X_n$} with *r* categorical attributes, $A_1, \ldots, A_r$ and the number of desired clusters, *k*, as input, we need (*r*+1)\**k* hash tables as our basic data structures. Actually, each hash table is the materialization of a histogram. The concept and structure of histogram has been discussed in Section 3.1. In the remaining parts of the paper, we will use histogram and hash table interchangeably.

As discussed in Section 3.2, each attribute $A_i$ determines an optimal partitioning $\lambda^{(i)}$ without considering other attributes. Storing $\lambda^{(i)}$ in its original format will be costly both in space and computation. Therefore, we only keep the histogram of $A_i$ on ***D***, denoted as $AH_i$, as the compact representation of $\lambda^{(i)}$. Since we have *r* attributes, *r* histograms are constructed.

Suppose that the partition of these *n* objects into *k* clusters is represented as a set of *k* sets of objects $C_l=\{l=1,\ldots,k\}$ or as a label vector $\tilde{\lambda} \in N^n$. For each $C_l$, we construct a histogram for each attribute separately. We denote the histogram of $A_i$ on $C_l$ as $CAH_{l,i}$. Hence, we need $r$ histograms for each $C_l$ and $r*k$ histograms for $\tilde{\lambda}$.

Overall, we need $(r+1)*k$ histograms totally.

**Example 1:** For example, Table 1 shows a categorical table with 10 records, each described by 2 categorical attributes. Only considering "*Attribute 1*", we can get the optimal partitioning {(1,2,5,7,10), (3,4,6,8,9)} with 2 clusters. Similarly, "*Attribute 2*" gives an optimal partitioning as {(1,4,9), (2,3,10), (5,6,7,8)} with 3 clusters. Suppose $k=2$, and $\tilde{\lambda}=\{(1,2,3,4,5), (6,7,8,9,10)\}$. The 6 histograms that we need are described in a vivid form, as shown in Fig. 2.

Table 1 Sample Categorical Data Set

| Record Number | Attribute 1 | Attribute 2 |
|---|---|---|
| 1 | M | A |
| 2 | M | B |
| 3 | F | B |
| 4 | F | A |
| 5 | M | C |
| 6 | F | C |
| 7 | M | C |
| 8 | F | C |
| 9 | F | A |
| 10 | M | B |

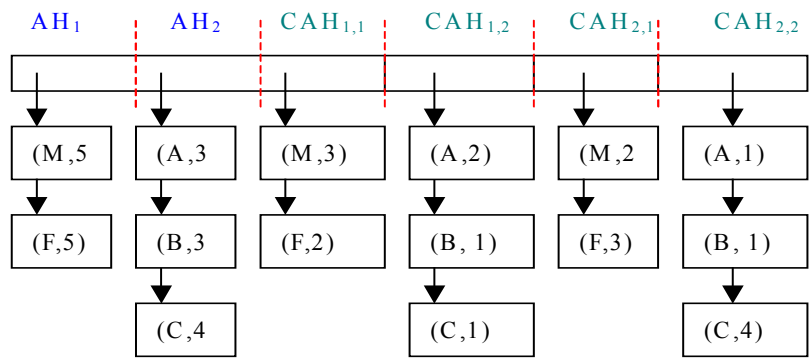

Fig. 2. The 6 Histograms for Example 1.

## 4.3 Computation of ANMI

In this section, we show how to use those histograms introduced in Section 4.2 to compute the ANMI value.

To compute the *ANMI* between a set of $r$ labelings, $\Lambda$, and a labeling $\tilde{\lambda}$, we only need to compute $\phi^{(NMI)}(\tilde{\lambda}, \lambda^{(i)})$ for each $\lambda^{(i)} \in \Lambda$. Therefore, we will focus on the computation of $\phi^{(NMI)}(\tilde{\lambda}, \lambda^{(i)})$. To be consistent with the description in Section 3.3, we use $\phi^{(NMI)}(\lambda^{(a)}, \lambda^{(b)})$ instead of $\phi^{(NMI)}(\tilde{\lambda}, \lambda^{(i)})$ for illustration by setting $\lambda^{(a)} = \tilde{\lambda}$ and $\lambda^{(b)} = \lambda^{(i)}$.

Recalling that $\phi^{(NMI)}(\lambda^{(a)}, \lambda^{(b)}) = \frac{2}{n}\sum_{h=1}^{k^{(a)}}\sum_{g=1}^{k^{(b)}} n_g^{(h)} \log_{k^{(a)} * k^{(b)}}(\frac{n_g^{(h)} n}{n^{(h)} n_g})$, where $n^{(h)}$ is the number of objects in cluster $C_h$ according to $\lambda^{(a)}$, and $n_g$ is the number of objects in cluster $C_g$ according to $\lambda^{(b)}$. $n_g^{(h)}$ is the number of objects in cluster $C_h$ according to $\lambda^{(a)}$ as well as in cluster $C_g$ according to $\lambda^{(b)}$. $k^{(a)}$ is the numbers of clusters in $\lambda^{(a)}$ and $k^{(b)}$ is the numbers of clusters in $\lambda^{(b)}$.

To compute the value of $\phi^{(NMI)}(\lambda^{(a)}, \lambda^{(b)})$, we must know 6 values, which are $n$, $k^{(a)}$, $k^{(b)}$, $n^{(h)}$, $n_g$ and $n_g^{(h)}$.

(1) For a given dataset, the value of $n$ is number of objects in the dataset and hence is fixed;

(2) Since $\lambda^{(a)} = \tilde{\lambda}$, so $k^{(a)} = k$;

(3) Since $\lambda^{(b)}$ is original partition derived from attribute $A_b$, hence $k^{(b)}$ is equal to the size of $AH_b$, where $AH_b$ is histogram of $A_b$ on **D**. Note that the value of $k^{(b)}$ is directly derived from the corresponding histogram.

(4) The value of $n^{(h)}$ is equal to the sum of frequencies of attribute values in the histogram $CAH_{h,i}$ for any $1 \leq i \leq r$.

(5) Suppose the cluster $C_g$ in $\lambda^{(b)}$ is determined by attribute value $v$, and then $n_g$ is equal to the frequency value of $v$ in histogram $AH_b$.

(6) As in (5), suppose the cluster $C_g$ in $\lambda^{(b)}$ is determined by attribute value $v$, and then $n_g^{(h)}$ is equal to the frequency value of $v$ in histogram $CAH_{h,b}$ if $v$ has an entry in $CAH_{h,b}$. otherwise, $n_g^{(h)} = 0$.

From (1) to (6), we know that the ANMI value can be computed by only using histograms and without accessing the original dataset. Thus, the computation is very effective.

**Example 2:** Continuing example 1, suppose we are trying to compute $\phi^{(NMI)}(\lambda^{(a)}, \lambda^{(b)})$, where $\lambda^{(a)} = \tilde{\lambda} = \{(1,2,3,4,5), (6,7,8,9,10)\}$ and $\lambda^{(b)} = \{(1,2,5,7,10), (3,4,6,8,9)\}$. In this case, we have $n=10$, $k^{(a)} = k = 2$, $\lambda^{(b)} = 2$. Further suppose that $C_h = (1,2,3,4,5)$ and $C_g = (1,2,5,7,10)$. We have $n^{(h)} = 3+2 = 5$ (according to $CAH_{1, 1}$), $n_g = 5$ (the value frequency of "M" in histogram $AH_{1,}$) and $n_g^{(h)} = 3$ (the value frequency of "M" in histogram $CAH_{1, 1}$).

### 4.4 The Algorithm

Fig.3 shows the $k$-ANMI algorithm. The collection of records is stored in a file on the disk and we read each record $t$ in sequence.

```
Algorithm k-ANMI
Input:    D    // the categorical database
          k    // the number of desired clusters
Output:   clusterings of D

/* Phase 1-initialization */
01  Begin
02      foreach transaction t in D
03          counter++
04          update histograms for each attribute
05          if counter<=k then
06              put t into cluster C_i with the number of counter
07          else
08              put t into cluster C_i with which has the smallest distance
09          write <t, i>

/* Phase 2-Iteration */
10      Repeat
11          not_moved =true
12          while not end of the database do
13              read next record < t, C_i >
14              moving t to an existing cluster , C_j to maximize ANMI
15              if C_i != C_j then
16                  write <t, j>
17                  not_moved =false
18      Until not_moved
19  End
```

**Fig. 3**. The $k$-ANMI algorithm.

In the initialization phase of the *k*-ANMI algorithm, we firstly select the first *k* records from the data set to construct initial histograms for each cluster. Each consequent record is put into the closed cluster according to Equation (6). The cluster label of each record is stored. At the same time, the histogram of partition derived from each attribute is also constructed and updated.

In iteration phase, we read each record *t* (in the same order as in initialization phase), move *t* to an existing cluster (possibly stay where it is) to maximize ANMI. After each move, the cluster identifier is updated. If no record is moved in one pass of all records, iteration phase terminates; otherwise, a new pass begins. Essentially, at each step we locally optimize the criterion ANMI. The key step is finding the destination cluster for moving a record according to the value of ANMI. How to efficiently compute ANMI using histograms has been discussed in Section 4.3.

### 4.5 Time and Space Complexities

**Worst-case analysis:** The time and space complexities of the *k*-ANMI algorithm depend on the size of dataset (*n*), the number of attributes (*r*), the number of the histograms, the size of every histogram, the number of clusters (*k*) and the iteration times (*I*).

To simplify the analysis, we will assume that every attribute has the same number of distinct attributes values, *p*. Then, in the worst case, in the initialization phase, the time complexity is $O(n*k*r*p)$. In the iteration phase, since the computation of ANMI requires $O(r*p^2*k)$ and hence this phase has time complexity $O(I*n*k^2*r*p^2)$. Totally, the algorithm has time complexity $O(I*n*k^2*r*p^2)$ in worst case.

The algorithm only needs to store $(r+1)*k$ histograms and the dataset in main memory, so the space complexity of our algorithm is $O(r*k*p+n*r)$.

**Practical analysis:** As pointed out in [10], categorical attributes usually have *small* domains. Typical categorical attributes domains considered for clustering consist of less than a hundred or, rarely, a thousand attribute values. An important of implication of the compactness of categorical domains is that the parameter, *p*, can be regarded to be very small. And the use of hashing technique in histograms also reduces the impact of *p*. So, in practice, the time complexity of *k*-ANMI can be expected to be $O(I*n*k^2*r*p)$.

The above analysis shows that the time complexity of *k*-ANMI is linear to the size of dataset, the number of attributes and the iteration times, which make this algorithm deserve good scalability.

## 5. Experimental Results

A performance study has been conducted to evaluate our method. In this section, we describe those experiments and the results. We ran our algorithm on real-life datasets obtained from the UCI Machine Learning Repository [25] to test its clustering performance against other algorithms.

### 5.1 Real Life Datasets and Evaluation Method

We experimented with three real-life datasets: the Congressional Votes dataset, the Mushroom dataset and the Wisconsin Breast Cancer dataset, which were obtained from the UCI

Machine Learning Repository [25]. Now we will give a brief introduction about these datasets.

- ✓ **Congressional Votes:** It is the United States Congressional Voting Records in 1984. Each record represents one Congressman's votes on 16 issues. All attributes are Boolean with Yes (denoted as *y*) and No (denoted as *n*) values. A classification label of Republican or Democrat is provided with each record. The dataset contains 435 records with 168 Republicans and 267 Democrats.
- ✓ **The Mushroom Dataset:** It has 22 attributes and 8124 records. Each record represents physical characteristics of a single mushroom. A classification label of poisonous or edible is provided with each record. The numbers of edible and poisonous mushrooms in the dataset are 4208 and 3916, respectively.
- ✓ **Wisconsin Breast Cancer Data[1]:** It has 699 instances with 9 attributes. Each record is labeled as *benign* (458 or 65.5%) or *malignant* (241 or 34.5%). In our literature, all attributes are considered categorical with values 1,2, …, 10.

Validating clustering results is a non-trivial task. In the presence of true labels, as in the case of the data sets we used, the clustering accuracy for measuring the clustering results was computed as follows. Given the final number of clusters, *k*, clustering accuracy *r* was defined as: $r = \frac{\sum_{i=1}^{k} a_i}{n}$, where *n* is the number of records in the dataset, $a_i$ is the number of instances occurring in both cluster *i* and its corresponding class, which had the maximal value. In other words, $a_i$ is the number of records with the class label that dominates cluster *i*. Consequently, the clustering error is defined as $e = 1 - r$.

## 5.2 Experiment Design

We studied the clustering found by four algorithms, our *k*-ANMI algorithm, the *Squeezer* algorithm introduced in [16], the GAClust algorithm proposed in [20] and ccdByEnsemble algorithm in [21].

Until now, there is no well-recognized standard methodology for categorical data clustering experiments. However, we observed that most clustering algorithms require the number of clusters as an input parameter, so in our experiments, we cluster each dataset into different number of clusters, varying from 2 to 9. For each fixed number of clusters, the clustering errors of different algorithms were compared.

In all the experiments, except for the number of clusters, all the parameters required by the *ccdByEnsemble* algorithm are set to be default as in [21]. The *Squeezer* algorithm requires *only* a similarity threshold as input parameter, so we set this parameter to a proper value to get the desired number of clusters. For the GAClust algorithm, we set the population size to be 50, and set other parameters to their default values[2].

---

[1] We use a dataset that is slightly different from its original format in UCI Machine Learning Repository, which has 683 instances with 444 benign records and 239 *malignant* records. It is public available at: http://research.cmis.csiro.au/rohanb/outliers/breast-cancer/brcancerall.dat.

[2] The source codes for GAClust are public available at: http://www.cs.umb.edu/~dana/GAClust/index.html. The readers may refer to this site for details about other parameters.

Moreover, since the clustering results of *k*-ANMI algorithm, *ccdByEnsemble* algorithm and *Squeezer* algorithm are fixed for a particular dataset when the parameters are fixed, only one run is used in the two algorithms. The GAClust algorithm is a genetic algorithm, so its outputs will differ in different runs. However, we observed in the experiments that the clustering error is very stable, so the clustering error of this algorithm is reported with its first run. In summary, we use one run to get the clustering errors for all the four algorithms.

### 5.3  Clustering Results on Congressional Voting (votes) Data

Fig. 4 shows the results on the *votes* dataset of different clustering algorithms. From Fig. 4, we can summarize the relative performance of these algorithms as Table 2. In Table 2, the numbers in column labelled by *k* (*k*=1, 2, 3 or 4) are the times that an algorithm has rank *k* among the four algorithms. For instance, in the 8 experiments, *Squeezer* algorithm performed second best in 2 cases, that is, it is ranked 2 for 2 times.

Compared to the other three algorithms, the *k*-ANMI algorithm performed best in most cases and never performed the worst. And the average clustering error of the *k*-ANMI algorithm was significantly smaller than that of other algorithms. Thus, the clustering performance of *k*-ANMI on the *votes* dataset is superior to all other three algorithms.

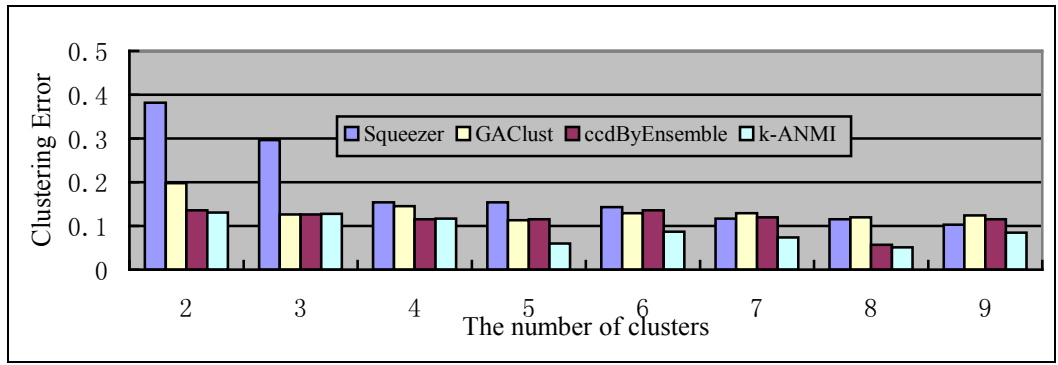

**Fig.4.** Clustering error vs. Different number of clusters (*votes* dataset)

**Table 2:** Relative performance of different clustering algorithms (*votes* dataset)

| Ranking | 1 | 2 | 3 | 4 | Average Clustering Error |
|---|---|---|---|---|---|
| *Squeezer* | 0 | 2 | 2 | 4 | 0.163 |
| GAClust | 1 | 2 | 2 | 3 | 0.136 |
| *ccdByEnsemble* | 2 | 2 | 4 | 0 | 0.115 |
| *k*-ANMI | 6 | 2 | 0 | 0 | **0.092** |

### 5.4  Clustering Results on Mushroom Data

The experimental results on the *mushroom* dataset are described in Fig. 5 and the relative performance is summarized in Table 3. As shown in Fig. 5 and Table 3, our algorithm beats all the other algorithms in average clustering error. Furthermore, although the *k*-ANMI algorithm didn't always perform best on this dataset, it performed best in 5 cases and never performed worst. That is, *k*-ANMI algorithm performed best in majority of the cases.

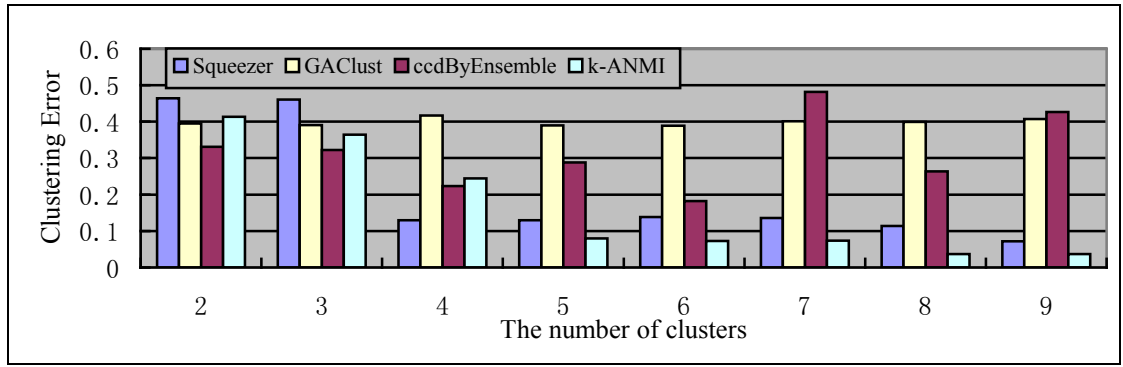

**Fig.5.** Clustering error vs. Different number of clusters (mushroom dataset)

**Table 3:** Relative performance of different clustering algorithms (mushroom dataset)

| Ranking | 1 | 2 | 3 | 4 | Average Clustering Error |
|---|---|---|---|---|---|
| *Squeezer* | 1 | 5 | 0 | 2 | 0.206 |
| **GAClust** | 0 | 1 | 3 | 4 | 0.393 |
| *ccdByEnsemble* | 2 | 1 | 3 | 2 | 0.315 |
| ***k*-ANMI** | 5 | 1 | 2 | 0 | **0.165** |

Moreover, the results of *k*-ANMI algorithm are significantly better than that of *ccdByEnsemble* algorithm in most cases. It demonstrates that direct optimization strategy utilized in *k*-ANMI is more desirable than the intuitive heuristics in *ccdByEnsemble* algorithm.

### 5.5 Clustering Results on Cancer Data

The experimental results on the *cancer* dataset are described in Fig. 6 and the summarization on the relative performance of the 4 algorithms is given in Table 4. From Fig. 6 and Table 4, some important observations are summarized as follows:

(1) Our algorithm beats all the other algorithms with respect to average clustering error.

(2) The *k*-ANMI algorithm almost performed best in all cases (except for the case when the number of clusters is 5); Furthermore, in almost every case, *k*-ANMI algorithm achieves better output than that of *ccdByEnsemble* algorithm, which verify the effectiveness of the direct optimization strategy in *k*-ANMI. In particular, when the number of clusters is set to 2 (the true number of clusters for the *cancer* dataset), our *k*-ANMI algorithm is able to get clustering output whose clustering error is significantly smaller than that of other algorithms.

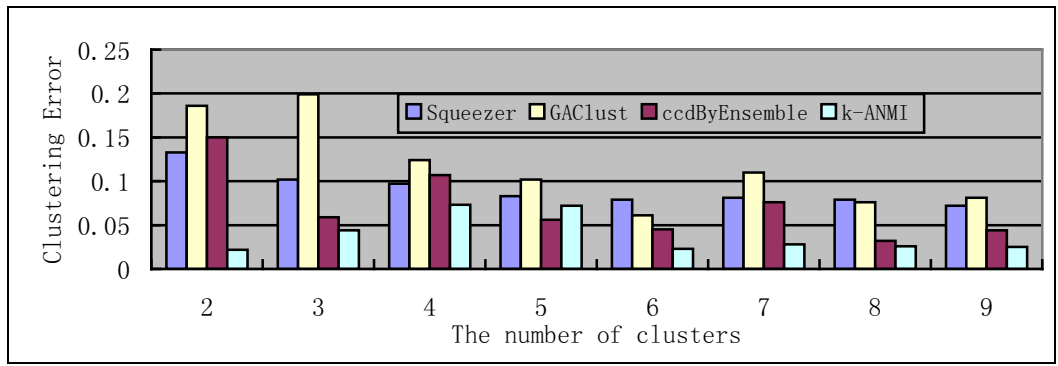

**Fig.6.** Clustering error vs. Different number of clusters (cancer dataset)

**Table 4:** Relative performance of different clustering algorithms (cancer dataset)

| Ranking | 1 | 2 | 3 | 4 | Average Clustering Error |
|---|---|---|---|---|---|
| *Squeezer* | 0 | 1 | 5 | 2 | 0.091 |
| GAClust | 0 | 0 | 2 | 6 | 0.117 |
| *ccdByEnsemble* | 1 | 6 | 1 | 0 | 0.071 |
| *k*-ANMI | 7 | 1 | 0 | 0 | **0.039** |

## 5.6 Scalable Tests

The purpose of this experiment was to test the scalability of the *k*-ANMI algorithm when handling very large datasets. A synthesized categorical dataset created with the software developed by Dana Cristofor [20] is used. The data size (i.e., number of rows), the number of attributes and the number of classes are the major parameters in the synthesized categorical data generation, which were set to be 100,000, 10 and 10 separately. Moreover, we set the random generator seed to 5. We will refer to this synthesized dataset with name of DS1.

We tested two types of scalability of the *k*-ANMI algorithm on large dataset. The first one is the scalability against the number of objects for a given number of clusters and the second is the scalability against the number of clusters for a given number of objects. Our *k*-ANMI algorithm was implemented in Java. All experiments were conducted on a Pentium4-2.4G machine with 512 M of RAM and running Windows 2000. Fig. 7 shows the results of using *k*-ANMI to find 2 clusters with different number of objects. Fig. 8 shows the results of using *k*-ANMI to find different number of clusters on DS1 dataset.

One important observation from these figures was that the run time of *k*-ANMI algorithm tends to increase linearly as the number of records is increased, which is highly desired in real data mining applications.

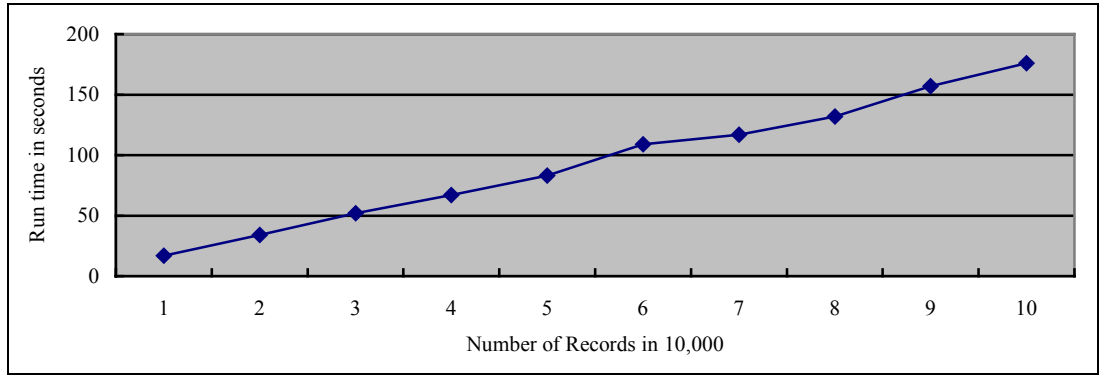

**Fig. 7**. Scalability of *k*-ANMI to the number of objects when mining 2 clusters from DS1 dataset

Furthermore, although the run time of *k*-ANMI algorithm doesn't increase linearly as the number of clusters is increased, it at least achieves good scalability at an acceptable level.

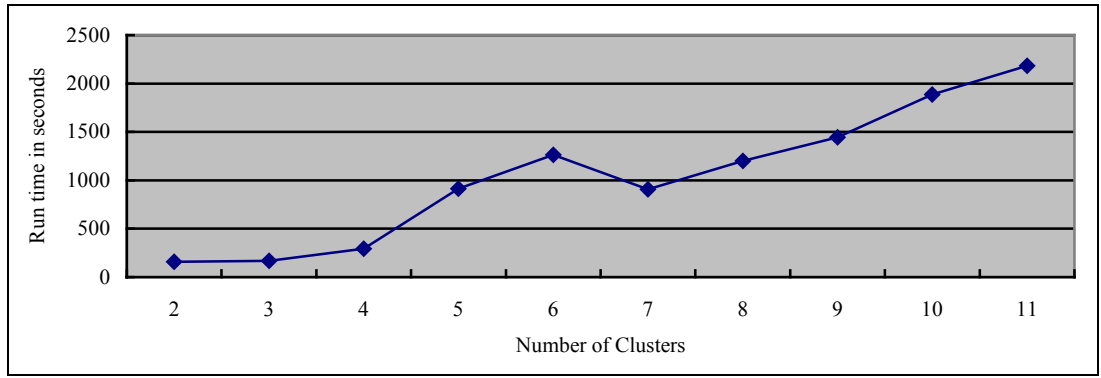

**Fig. 8**. Scalability of *k*-ANMI to the number of clusters when clustering 100,000 records of the DS1 dataset

## 6. Conclusions

In this paper, we propose a new *k*-means like clustering algorithm called *k*-ANMI for categorical data, which tries to directly optimize a mutual information sharing based object function. Empirical evidences show that our method is effective in practice.

For future work, we are planning to design fast genetic clustering algorithms for categorical data using ANMI.

## Acknowledgements

This work was supported by The High Technology Research and Development Program of China (Grant No. 2003AA4Z3370, Grant No. 2003AA413021), the National Nature Science Foundation of China (Grant No. 40301038) and the IBM SUR Research Fund.